\documentclass{article}
\usepackage{spconf,amsmath,graphicx}

\usepackage{microtype}
\usepackage{graphicx}
\usepackage{booktabs} % for professional tables
\usepackage{hyperref}
\usepackage[utf8]{inputenc} % allow utf-8 input
\usepackage[T1]{fontenc}    % use 8-bit T1 fonts
\usepackage{hyperref}       % hyperlinks
\usepackage{url}            % simple URL typesetting
\usepackage{booktabs}       % professional-quality tables
\usepackage{amsfonts}       % blackboard math symbols
\usepackage{nicefrac}       % compact symbols for 1/2, etc.
\usepackage{microtype}      % microtypography

\usepackage{times}
\usepackage{latexsym}
\usepackage{enumitem}

\usepackage{graphicx}
\usepackage{amssymb}
\usepackage{comment}

\usepackage{caption}
\usepackage{subcaption}
\usepackage{booktabs}

\usepackage{multicol}
\usepackage{multirow}
\usepackage{array, tabularx,  ragged2e,  booktabs}
\usepackage[numbers]{natbib}

\usepackage[utf8]{inputenc}
\usepackage{graphicx}
\usepackage{pgfplots}
\pgfplotsset{compat=1.14}

\usepackage{ctable}
\usepackage{wrapfig}
\usepackage{tikz}
\usetikzlibrary{positioning,arrows,trees,shapes,fit,shadows}

\usepackage{amsfonts}
\usepackage{ifthen}
\usepackage{booktabs}
\usepackage{amssymb}
\usepackage{supertabular}
\usepackage{array}
\usepackage{multirow}
\usepackage{fancyhdr}
\usepackage[normalem]{ulem}

\usepackage{amssymb}
\usepackage{supertabular}
\usepackage{array}
\usepackage{multirow}
\usepackage{fancyhdr}
\usepackage{psfrag}
\usepackage{amsmath}
\usepackage{hhline}
\usepackage{type1cm}
\usepackage{lettrine}
\usepackage[colorinlistoftodos,prependcaption,textsize=tiny]{todonotes}

\usepackage{algorithm}
\usepackage{algorithmic} % enabling algorithmic make the algo not working

\usepackage{xspace}

\usepackage{babel}
\usepackage[export]{adjustbox}
\usepackage{cleveref}
\usepackage{scrextend}

\crefformat{section}{\S#2#1#3} % see manual of cleveref, section 8.2.1
\crefformat{subsection}{\S#2#1#3}
\crefformat{subsubsection}{\S#2#1#3}

\usepackage{amsmath,amsfonts,bm}

\def\eqref#1{equation~\ref{#1}}
\def\1{\bm{1}}

\makeatletter
\newcommand{\xRightarrow}[2][]{\ext@arrow 0359\Rightarrowfill@{#1}{#2}}
\makeatother

\DeclareMathAlphabet{\mathsfit}{\encodingdefault}{\sfdefault}{m}{sl}
\SetMathAlphabet{\mathsfit}{bold}{\encodingdefault}{\sfdefault}{bx}{n}

\newcommand{\eg}{{\em e.g.,}\xspace}

\newcommand{\Ni}{({\em i})~}
\newcommand{\Nii}{({\em ii})~}
\newcommand{\Niii}{({\em iii})~}
\newcommand{\Niv}{({\em iv})~}

\newcommand{\textaug}{\textit{Text-aug}\xspace}
\newcommand{\effectsaug}{\textit{Effects-aug}\xspace}

\newcommand{\covost}{{CoVoST-2}\xspace}
\newcommand{\epst}{{Europarl-ST}\xspace}
\newcommand{\mtedx}{{mTEDx}\xspace}
\newcommand{\mustc}{{MuST-C}\xspace}

\usepackage{float}

\DeclareCaptionLabelFormat{andtable}{#1~#2  \&  \tablename~\thetable}

\usepackage{tikz-cd}
\usepackage{mathtools}
\usepackage{todonotes}
\setcitestyle{square}

\title{Improving Speech-to-Speech Translation Through Unlabeled Text}
\name{Xuan-Phi Nguyen$^{* \dagger}$\thanks{$^{*}$Work done during an internship at Meta AI.}, Sravya Popuri$^{\star}$, Changhan Wang$^{\star}$, Yun Tang$^{\star}$, Ilia Kulikov$^{\star}$ and Hongyu Gong$^{\star}$}
\address{$^{\star}$ Meta AI, USA \\
      $^{\dagger}$ Nanyang Technological University, Singapore\\
      \texttt{nguyenxu002@e.ntu.edu.sg}\\\texttt{\{spopuri,changhan,yuntang,kulikov,hygong\}@meta.com} \\}
\begin{document}
%\ninept
%
\maketitle
\begin{abstract}
Direct speech-to-speech translation (S2ST) is among the most challenging problems in the translation paradigm due to the significant scarcity of S2ST data. While effort has been made to increase the data size from unlabeled speech by cascading pretrained speech recognition (ASR), machine translation (MT) and text-to-speech (TTS) models; unlabeled text has remained relatively under-utilized to improve S2ST. We propose an effective way to utilize the massive existing unlabeled text from different languages to create a large amount of S2ST data to improve S2ST performance by applying various acoustic effects to the generated synthetic data. Empirically our method outperforms the state of the art in Spanish-English translation by up to $2$ BLEU. Significant gains by the proposed method are demonstrated in extremely low-resource settings for both Spanish-English and Russian-English translations.
\end{abstract}
\begin{keywords}
Speech-to-speech translation, augmentation, unlabeled text
\end{keywords}

\section{Introduction}
\label{sec:intro}

Translating speech of a language to speech of another can be done by trivially cascading an automatic speech recognition (ASR) \cite{self_train_asr_kahn2020}, machine translation (MT) \cite{vaswani2017attention,criss2020,swavumt2021}, or combinatory speech-to-text translation (S2T) \cite{s2t_bansal2017towards} and finally a text-to-speech (TTS) systems \cite{vits_kim2021conditional,fastspeech_ren2019,janus_iii_lavie1997janus,ATR_speech_to_speech_1597243}. But such process suffers from significant inference latency and is prone to error propagation through each stage. Alternatively, there is a growing interest in developing direct speech-to-speech translation systems (S2ST) \cite{direct_s2s_jia2019,s2s_discrete_units_lee2021,enhanced_s2u_popuri2022}. Not only do these systems have faster inference but also allow translations between unwritten languages and dialects \cite{textless-s2s-lee-etal-2022}. Recent speech-to-speech model \cite{enhanced_s2u_popuri2022} employs a self-supervised pretrained speech encoder \cite{wav2vec2_baevski2020} and a discrete unit mBART model to train a speech-to-unit translation (S2UT), where the target speech is converted into discrete units \cite{lakhotia2021generative}. Such units are trained with self-supervision to group speech frames based on their linguistic and prosodic information using $k$-means clustering \cite{discrete_unit_polyak2021speech}. 
% \alert{discrete units to speech}

% \hongyu{what does self-supervised mean here? SSL pretrained encoder?} --> yes

Despite promising capability, direct speech-to-speech (S2ST) models suffer from significant data scarcity due to the challenge of collecting human annotated parallel speech. To improve S2ST performance, apart from self-supervised pretraining on unlabeled speech \cite{direct_s2s_jia2019,wav2vec2_baevski2020,mbart_liu2020multilingual}, \cite{enhanced_s2u_popuri2022} also tried to generate extra synthetic S2ST data from speech recognition (ASR) data by converting its texts to speech in the target language using a cascaded MT-TTS system \citep{weaklysup_s2t_jia2019leveraging}. Nonetheless, such work only makes use of the available audio data, while not utilizing the existing unlabeled text data from numerous languages, sources and domains \cite{ccnet_data_wenzek2019,mbart_liu2020multilingual,xlm_conneau2019cross}. Text data is known to be much more massive and diverse than the current speech data. However, such textual data may be difficult to deal with in the speech paradigm as they lack crucial information about speakers, speed, pitch and emotions.

In this work, we present an effective strategy to generate synthetic speech-to-speech training data from the abundant unlabeled text data so that the resulting speech data is not only diverse in terms of semantic content, but also randomly varying in acoustic features such as speaker tones. Our approach consists of two processes: \Ni ``\textaug'' data generation, which is the synthetic S2ST data created from unlabeled text; and \Nii ``\effectsaug'' process, which is an on-the-fly speech augmentation process that transforms toneless \textaug speech into a varying-tone and noisy version that tries to mimic the distribution of real speech data. Our approach does not introduce any extra model or data supervision besides those used in the recent S2UT baseline \cite{enhanced_s2u_popuri2022}, which use supervised MT and TTS models. Instead, we utilize an unsupervised MT model \cite{criss2020} to generate data. In the experiments, our method achieves up to 35.2 BLEU on the \covost Es-En task and 35.1 BLEU on \epst En-Es task, surpassing the state-of-the-art approach \cite{enhanced_s2u_popuri2022} by up to 2 BLEU. Further analysis shows that \effectsaug is a crucial step for the extra data to improve the performance. We also demonstrate that our method achieves significant performance gain of up to 28 BLEU in low-resource speech-to-speech setups with only 10 hours (hr) S2ST data.

\section{Background}
\label{sec:background}

Success in self-supervised training of speech encoder \cite{wav2vec2_baevski2020} enables significant advancements in various speech processing tasks, ranging from ASR \citep{self_train_asr_kahn2020,hubert_hsu2021}, speech-to-text translation (S2T) \cite{s2t_bansal2017towards,s2t_self_learn_wang2021large}, which are critical components in building speech-to-speech translation systems \cite{s2s_discrete_units_lee2021,enhanced_s2u_popuri2022}. Such pretrained encoder produces speech hidden representations that can be discretized into units that condense semantic and prosodic information \cite{hubert_hsu2021}. Specifically, the self-supervised HuBERT model \cite{hubert_hsu2021} is trained to encode input speech into discrete units by performing $k$-means clustering over the hidden vectors by a pretrained $k$-means model.

\begin{figure}
    \begin{center}
    \centerline{\includegraphics[width=0.3\columnwidth]{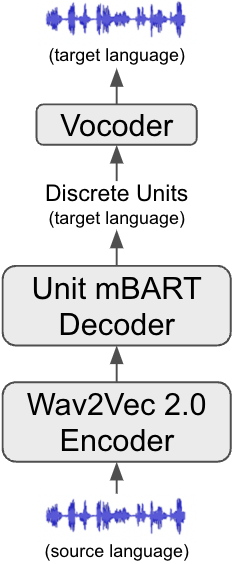}}
    \caption{Architecture of Speech-to-Unit translation and Unit vocoder models.}
    \label{fig:s2u_model}
    \vspace{-1em}
    \end{center}
    % \vskip -0.2in
    \vspace{-1.2em}
\end{figure}

Recent S2ST models \cite{direct_s2s_jia2019,textless-s2s-lee-etal-2022,enhanced_s2u_popuri2022} make use of such $k$-means units that, instead of directly generating an audio signal, which is considerably slow, they generate shorter unit sequences with a heavy speech-to-unit translation (S2UT) model and use a lightweight vocoder \cite{hifigan_vocoder_polyak2021} to convert units to output audios. Specifically, the S2UT model is an attention-based Seq2Seq model \cite{vaswani2017attention}. Its encoder is a  Wav2vec 2.0 \cite{wav2vec2_baevski2020} model that is pretrained to encode speech representations from unlabeled audios. It consists of a multi-layer convolutional network to encode raw audio signal, followed by a Transformer \cite{vaswani2017attention} (or Conformer \cite{conform_gulati2020}) encoder to produce contextual representations for the audio. Meanwhile, the decoder is a unit-mBART \cite{mbart_liu2020multilingual}, which is pretrained with masked language modeling on the unsupervised reduced discrete unit data derived from unlabeled speech via the HuBERT-$k$-means model \cite{hubert_hsu2021}. During S2S training, the S2UT model is initialized with the pretrained models \cite{wav2vec2_baevski2020,mbart_liu2020multilingual} and then finetuned with speech-to-unit data, which is in fact the S2S data where the target speech is converted to discrete units. During finetuning, Popuri et al \cite{enhanced_s2u_popuri2022} suggest that it is most beneficial to freeze the decoder parameters, except its layer-norm layers \cite{layernorm_ba2016layer} in the attention modules. \Cref{fig:s2u_model} depicts the architecture of the direct speech-to-speech translation system. More importantly, Popuri et al \cite{enhanced_s2u_popuri2022} also make use of intensive data augmentation, where extra supervised speech recognition (ASR) data is used with pretrained MT \cite{vaswani2017attention} and TTS \cite{fastspeech_ren2019} models to synthetically generate more speech-to-speech data for training, which profoundly improves the performance. Our approach is built on top of this work in that, in addition to speech-based data augmentation, we introduce an effective way to convert the existing massive unlabeled text data into prosodically diverse speech-to-speech data to add into the training data pool.

\section{Method}
\label{sec:method}

This section describes our strategy in building extra training data from unlabeled text (\cref{ssec:text_aug_creation}) and the training process that augments such synthetic data to be acoustically diverse (\cref{ssec:effects_aug_augmentation}).

\vspace{-0.5em}
\subsection{Text-Aug Data Creation}
\label{ssec:text_aug_creation}

\begin{figure}
    \begin{center}
    \centerline{\includegraphics[width=0.9\columnwidth]{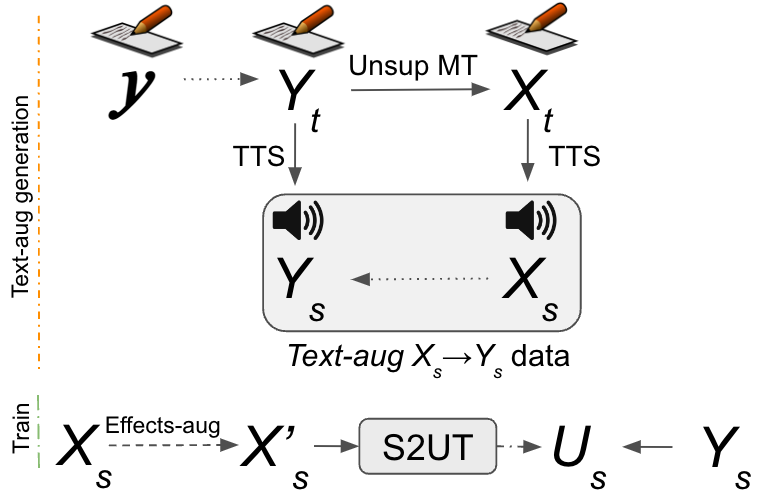}}
    \caption{Generation process of \textaug data for $X_s\rightarrow Y_s$ and the S2U training process with \effectsaug augmentation.}
    \label{fig:text_aug_creation}
    \vspace{-0.8em}
    \end{center}
    % \vskip -0.2in
    \vspace{-1.2em}
\end{figure}

\textaug data is the extra synthetic S2ST dataset that is created from unlabeled text. \Cref{fig:text_aug_creation} illustrates how it is generated and trained. Specifically, given an unlabeled text corpus $\mathcal{Y}$ of the target language, we first clean them and remove non-conversational sentences, such as those that contain URLs, special characters, words in brackets, etc, to obtain a cleaner unlabeled text corpus $Y_t$. We then use an unsupervised MT model, namely CRISS \cite{criss2020}, to translate $Y_t$ into $X_t$ text corpus of the source language. The resulting data will undergo further filtering to remove pairs with excessive length difference or repetitions. Then, both source ($X_t$) and target ($Y_t$) texts are converted into speech $X_s$ and $Y_s$ using existing text-to-speech (TTS) models \cite{vits_kim2021conditional}. The target speech $Y_s$ is then transformed into reduced units $U_s$ to form the speech-to-unit training data.

\subsection{Effects-Aug Data Augmentation}
\label{ssec:effects_aug_augmentation}

As the \textaug speech is generated by a single TTS model, which was trained from single-speaker data, its speech prosody is largely monotonic, single-style, unnatural. Training the S2UT model with a large amount of such data will overwhelm the existing real speech data and cause a shift in the real data distribution at convergence, thus leading to failure in generalization in test sets. Therefore, during training, we apply a series of random acoustic effects and background noise impositions, which we collectively call \effectsaug, to each \textaug speech utterance of the source language. 
Specifically, for each input utterance $X$, we augment $X$ into $\bar{X}$ as follows:
\vspace{-0.2em}
\begin{equation}
    \bar{X} = f_1(f_2(...f_n(X)...))
\end{equation}
where each function $f_i$ will, at $p\%$ chance, either apply augmentation $h_i$ to input $x$ with a random configuration or do nothing and return $x$. Each $h_i \in \{h_1,...,h_n\}$ represents a distinct type of audio perturbation effects, which include \href{https://sox.sourceforge.net/}{SoX effects} like speed, pitch and frequency variations, low-pass filter; as well as random mixing with real environmental noise or music segments \cite{musan2015} at different positions and signal-to-noise (SNR) ratio \cite{snr_plapous2006improved}. We argue that applying multiple random \effectsaug augmentations on the unnatural \textaug data may, to a certain degree, help them mimic the realistic conditions, where there can be multiple speakers with different tones, pitches and emotions recording their voices in various environments such as coffee shops or construction sites. Our analysis experiments show that without such augmentation processes, the \textaug data actually cause performance decline instead, as they may be unrealistic that they potentially cause a shift in the convergence point of the model.

% old stuff
% Specifically, using the \href{https://sox.sourceforge.net/}{SoX audio processing tool}, a synthetic input speech utterance will be augmented with a series of various audio effects , such as faster or slower speed, changes in pitch and frequency, low-pass filter, etc. Each effect type will have 50\% chance being applied with a random configuration, such as increase speed by 1.1 ratio. In addition, background noise segments, which can contain environmental noise or music \cite{musan2015}, will also be randomly chosen and imposed on the input speech at random position and signal-to-noise ratio \cite{snr_plapous2006improved}. We argue that applying multiple random \effectsaug augmentations on the unnatural \textaug data may, to a certain degree, help them mimic the realistic speech conditions, where there can be multiple speakers with different tones, pitches and emotions recording their voices in various environments such as coffee shops, bars, or construction sites. Our analysis experiments show that without such augmentation processes, the \textaug data actually cause performance decline instead, as they may be unrealistic that they potentially cause a shift in the convergence point of the model.

\begin{table*}[t]
% \begin{table}[h]
    \begin{center}
    \vspace{-1em}
    \caption{ASR-BLEU scores for Spanish (Es) - English (En) speech-to-speech translation tasks.
    }
    % \vspace{-0.5em}
    % \resizebox{0.8\textwidth}{!}{
    % \setlength{\tabcolsep}{1.5pt}
    \begin{tabular}{l|cccc|ccc}
    \toprule
    \multirow{2}{*}{\bf Method} & \multicolumn{4}{c}{\bf Es-En} & \multicolumn{3}{c}{\bf En-Es} \\
                    % \cmidrule(lr){2-9}\cmidrule(lr){10-15}  \cmidrule(lr){16-17}
                    & {\bf CoVoST-2} & {\bf Europarl-ST} & {\bf mTEDx} & {\bf Avg} & {\bf Europarl-ST} & {\bf MuST-C} & {\bf Avg} \\
    \midrule
    S2T (w2v2-L) + TTS & 28.4 &	23.6	& 21.5	& 24.5	& 32.6	& 30.1	& 31.4 \\
    ASR + MT + TTS	   & 33.8	& 29.1	& 32.4	& 31.8	& 28.8	& 34.2	& 31.5\\
    S2UT  \cite{s2s_discrete_units_lee2021}      &  22.7 & 18.0 & 16.9  & 19.2 & 25.8 &  24.3  & 25.1 \\
    \midrule
    S2UT + ASR aug \cite{enhanced_s2u_popuri2022} 	& 33.5	& 28.6	& 29.1	& 30.4	& 33.6	& 33.7	& 33.7\\
    \cite{enhanced_s2u_popuri2022} + Back-translation \cite{understand_bt_edunov2018}	& 34.3	& 30.3	& 30.1	& 31.6	& 33.9	& 33.7	& 33.8\\
    \midrule
    % \cite{enhanced_s2u_popuri2022} + w/ \effectsaug	 & 33.7	& 30.4	& 30.0	& 31.4	& 33.9	& 33.8	& 33.9 \\
    \cite{enhanced_s2u_popuri2022} + \textaug + \effectsaug	& 35.1	& 31.1	& 31.0	& 32.4	& 35.1	& 34.1	& 34.6\\
    \bottomrule
    \end{tabular}
    % }
    \vspace{-0.5em}
    \label{table:esen}
    \end{center}
\end{table*}

\section{Experiments}
\label{sec:experiments}

\subsection{Spanish-English Speech-to-Speech Translation}
\label{ssec:main_esen_exp}

% \hongyu{A high-level comment: shall we have a standalone paragraph ``Training Data'' to describe data (real data + weakly supervised data) used by previous works in a centralized way? This could provide more clarity to readers if our paper is self-contained so that they don't need to look up data configs in references.}

% \subsection{Training Data}\label{ssec:training_data}
Regarding training data, we reuse the same S2ST dataset as used by Popuri et al \cite{enhanced_s2u_popuri2022}. Specifically, we use single-speaker TTS models to convert the target text of S2T data from various sources, namely \covost \cite{covost2_wang2020covost}, \epst \cite{europarl_iranzo2020}, mTEDx \cite{mtedx_elizabeth2021multilingual}. These data is joint with VoxPopuli \cite{textless-s2s-lee-etal-2022} S2ST data. In addition, we also include the extra ASR augmentation data introduced in Popuri et al \cite{enhanced_s2u_popuri2022}, which originates from MLS \cite{mls_pratap2020mls}, CommonVoice \cite{common_voice_ardila2019}, Librispeech \cite{libri_kahn2020libri} and TEDLIUM \cite{tedlium_hernandez2018ted}. These sources result in the total original dataset sizes of 1800 and 2880 hours (hr) for Es$\rightarrow$En and En$\rightarrow$Es respectively.

In terms of model setup, similar to \cite{enhanced_s2u_popuri2022}, we use the multilingual HuBERT and $k$-means model \cite{textless-s2s-lee-etal-2022}, which was pretrained from unlabeled VoxPopuli speech data \cite{voxpopuli_wang2021}. The S2U model's encoder is a large Conformer Wav2Vec 2.0 \cite{wav2vec2_baevski2020,conform_gulati2020} pretrained with Libri-light \cite{libri_kahn2020libri} for En and VoxPopuli \cite{voxpopuli_wang2021} for Es. The decoder is the unit mBART \cite{mbart_liu2020multilingual} that was pretrained from reduced units derived from the aforementioned unlabeled speech with the HuBERT-$k$-means models. 

Regarding setups relating to our method, to produce the \textaug dataset, we use the pretrained unsupervised MT model CRISS \cite{criss2020} to translate $\sim$12M En and 12M Es monolingual sentences, which are randomly sampled from the CC25 \cite{ccnet_data_wenzek2019} corpora, into Es and En for respectively. 
After further filtering and TTS speech conversion \cite{vits_kim2021conditional}, we obtain $\sim$14K and 21K hours of audio \textaug data for Es$\rightarrow$En and En$\rightarrow$Es tasks, which are almost 10x the original data \cite{enhanced_s2u_popuri2022}. Despite its much larger size, during S2U finetuning, we sample the original and \textaug data at a 50:50 sampling ratio to ensure that the model has sufficient exposure to real audios to avoid further distribution shift. We compare our method  with the state of the art \cite{enhanced_s2u_popuri2022}, along with related baselines such as the cascaded S2T+TTS and ASR+MT+TTS systems or \cite{enhanced_s2u_popuri2022} with back-translation data from unlabeled speech \cite{understand_bt_edunov2018}. In terms of \effectsaug settings, we randomly select at $p=50\%$ chance: \Ni speed variations by 0.95-1.05 ratio, \Nii pitch variations by 0.95-1.05 ratio, \Niii low-pass filter with cut-off frequency in 300-1000Hz, \Niv up to 4 noise utterances chosen in the Musan corpus \cite{musan2015} at SNR ratio between 25-35. To stabilize the model, we average the best 10 checkpoints after training for 50K updates.

\Cref{table:esen} compares the ASR-BLEU \cite{enhanced_s2u_popuri2022} scores of our method against related baselines on the \covost, \epst and \mtedx Es$\rightarrow$En S2ST task; as well as the \epst and \mustc En$\rightarrow$Es task. Specifically, on average across multiple test sets, our method surpasses \cite{enhanced_s2u_popuri2022} by up to 2 BLEU for Es$\rightarrow$En and 1 BLEU for En$\rightarrow$Es tasks. In addition, despite achieving improvements with extra back-translation data \cite{understand_bt_edunov2018} generated from unlabeled speech \cite{voxpopuli_wang2021}, \cite{enhanced_s2u_popuri2022} still lags behind our method by 1 BLEU score on average.

\subsection{Low-resource and Unsupervised Translation}
\label{ssec:low_resource}

% \begin{table}[t]
\begin{table}[h]
    \begin{center}
    % \vspace{-0.5em}
    \caption{Averaged test scores for low-resource and unsupervised Es$\rightarrow$En and En$\rightarrow$Es S2ST tasks.
    }
    % \vspace{-0.5em}
    \resizebox{\columnwidth}{!}{
    \begin{tabular}{l|cc}
    \toprule
        % \multirow{2}{*}{\bf Method} & \multicolumn{4}{c}{\bf Es-En} & \multicolumn{3}{c}{\bf En-Es} \\
        %                 % \cmidrule(lr){2-9}\cmidrule(lr){10-15}  \cmidrule(lr){16-17}
        %                 & {\bf CoVoST-2} & {\bf Europarl-ST} & {\bf mTEDx} & {\bf Avg} & {\bf Europarl-ST} & {\bf MuST-C} & {\bf Avg} \\
        % % NOTE: 0hr unsupervised
        % 0hr + our method & 14.4	& 21.4	& 17.4	 & 17.7  & 19.4	& 16.8 & 18.1 \\
        % S2T-data 10hr	& 2.1	& 0.0	& 3.0	& 1.7	& 0.4	& 0.2	& 0.3 \\
        % \hspace{0.2em} + \textaug+\effectsaug	    & 30.1	& 25.5	& 26.4	& 27.3	& 29.3	& 27.4	& 28.4 \\
        % S2T-data 50hr	& 25.5	& 19.5	& 19.2	& 21.4	& 19.6	& 18.1	& 18.9\\
        % \hspace{0.2em} + \textaug+\effectsaug	& 32.3	& 27.7	& 27.7	& 29.2	& 31.3	& 29.2	& 30.3\\
        % S2T-data 100hr	& 29.6	& 24.9	& 24.1	& 26.2	& 26.6	& 25.6	& 26.1\\
        % \hspace{0.2em} + \textaug+\effectsaug	& 32.7	& 28.5	& 27.8	& 29.7	& 31.5	& 29.3	& 30.4\\
     & Es-En (Avg) & En-Es (Avg) \\
    \midrule
    \multicolumn{3}{l}{\bf Low-resource S2T data}\\
    \midrule
    % 10hr / Text-Effects-aug	& 1.7 / 27.3		& 0.3 \\
    10hr / 10hr+Text-Effects-aug	& 1.7 / 27.3		& 0.3 / 28.4 \\
    % \hspace{0.2em} + \textaug+\effectsaug	   & 27.3	& 28.4 \\
    50hr / 50hr+Text-Effects-aug	& 21.4 / 29.2	& 18.9 / 30.3\\
    % \hspace{0.2em} + \textaug+\effectsaug	& 29.2	& 30.3\\
    100hr / 100hr+Text-Effects-aug	& 26.2 / 29.7	& 26.1 / 30.4\\
    % \hspace{0.2em} + \textaug+\effectsaug	& 29.7 & 30.4\\
    \midrule
    \multicolumn{3}{l}{\bf Unsupervised}\\
    0hr + Text-Effects-aug  & 17.7 & 18.1 \\
    % ---
    % ---
    \bottomrule
    \end{tabular}
    }
    \vspace{-0.5em}
    \label{table:low_res}
    \end{center}
\end{table}

To demonstrate the effectiveness of the \textaug data, we evaluate our method in extremely low-resource conditions, where we use only 10hr, 50hr and 100hr of S2ST data randomly sampled from the original training data \cite{s2s_discrete_units_lee2021}. \Cref{table:low_res} compare the averaged test scores for Es-En and En-Es between each low-resource performance of S2UT models \cite{enhanced_s2u_popuri2022} trained with different original S2ST data (\eg\ 10hr) and those trained with both S2ST and our synthetic \textaug (\eg\ 10hr+Text-Effects-aug). The results indicate that our method offers significant performance boost for extremely low-resource conditions (10hr), while diminishing return is observed as the amount of S2ST data with natural speech input increase to 100hr. Furthermore, in the unsupervised setup without any natural-speech S2ST data, our method is able to achieve 17.1 and 18.1 BLEU for Es-En and En-Es respectively.

\subsection{Russian-to-English Translation}
\label{ssec:ruen}

\begin{table}[h]
    \begin{center}
    % \vspace{-0.5em}
    \caption{Scores on the CoVoST-2 Russian-English S2ST task.
    }
    % \vspace{-0.5em}
    % \resizebox{\columnwidth}{!}{
    % \setlength{\tabcolsep}{1.5pt}
    \begin{tabular}{l|c}
    \toprule
           & \covost Ru-En \\
% 	CoVoST-2
    \midrule
    10hr / 10hr+Text-Effects-aug 	& 0.1 / 28.1\\
     % + \textaug+\effectsaug	& 28.1\\
    25hr / 25hr+Text-Effects-aug   & 0.2 / 34.9\\
     % + \textaug+\effectsaug	& 34.9\\
    % ---
    \bottomrule
    \end{tabular}
    % }
    \vspace{-1em}
    \label{table:ruen}
    \end{center}
\end{table}

We also evaluate our method on Russian-English S2ST task. Specifically, we apply the same S2UT setup as \cite{enhanced_s2u_popuri2022} to train the model on a 10hr sub-sample and the full 25hr natural S2ST dataset for Ru-En task, which is produced from a combination of CoVoST-2 \cite{covost2_wang2020covost} and mTEDx \cite{mtedx_elizabeth2021multilingual} speech corpora. As it can be seen in \Cref{table:ruen}, our method achieves up to 34 BLEU gain compared to the standard S2ST setup, which is mostly attributed to the fact that Ru-En dataset is inherently extremely low-resource. The observation for Ru-En is thus consistent with that for low-resource Es-En explained in \cref{ssec:low_resource}.

\subsection{Ablation Studies}
\label{ssec:ablation_studies}

% \begin{table}[t]
\begin{table}[h]
    \begin{center}
    % \vspace{-0.5em}
    \caption{Ablation studies with average Es-En test scores that aim to analyze different aspects of our method.
    }
    % \vspace{-0.5em}
    \resizebox{\columnwidth}{!}{
    \begin{tabular}{l|c}
    \toprule
    Analysis       & Es-En (Avg) \\
    \midrule
    Baseline \cite{enhanced_s2u_popuri2022} & 30.4\\
    \midrule
    \multicolumn{2}{l}{\bf Which should kind of audio augmentation be applied ?}\\
    % \midrule
    \textaug only (no \effectsaug)   & 28.93\\
    \textaug + \effectsaug(no background noise)  & 31.97 \\
    \textaug + \effectsaug(no SoX effects) & 31.90\\
    % \textaug + \effectsaug & 32.43\\
    \midrule
    \multicolumn{2}{l}{\bf On which data should \effectsaug be applied?} \\
    % \midrule
    \textaug + \effectsaug[on \textaug data only] & 32.40\\
    \textaug + \effectsaug[on S2ST data only] & 30.53\\
    \textaug + \effectsaug[on S2ST \& \textaug data] & 32.43\\
    \midrule
    \multicolumn{2}{l}{\bf Should we apply \effectsaug on target audio too?} \\
    % \midrule
    % \textaug + \effectsaug[on source side] & 32.40\\
    \textaug + \effectsaug[on source \& target side] & 32.37\\
    \midrule
    % ----
    \multicolumn{2}{l}{\bf Should we omit \textaug entirely and use \effectsaug only?} \\
    \effectsaug[on S2ST data, no \textaug]	 & 31.40 \\
    \midrule
    \textaug + \effectsaug (Our proposed method) & 32.40\\
    % ---
    \bottomrule
    \end{tabular}
    }
    \vspace{-1.4em}
    \label{table:ablation}
    \end{center}
\end{table}

We conduct ablation study experiments to gain more insights into the method, which are presented in \Cref{table:ablation}. Specifically, we empirically answer the following questions by comparing the averaged test ASR-BLEU scores for Es$\rightarrow$En task.

\textbf{Which is the best audio augmentation?} It is shown that applying either SoX effects or background noises on the \textaug data brings in decent gain, while applying both yields the best results. However, skipping the \effectsaug suite causes a dramatic performance degradation due to the monotonicity and unnaturalness of these text-based synthetic data.

\textbf{On which data should \effectsaug be applied?} Applying \effectsaug on the \textaug data is shown to be critical, while augmenting the S2ST data as well gives a slight improvement.

\textbf{Should we apply \effectsaug on target audio too?} Meanwhile, applying augmentation on the target side audios is unnecessary as it is shown to make little difference.

\textbf{Should we omit Text-aug entirely and use Effects-aug only?} Without the synthetic \textaug data, applying \effectsaug augmentation on the original S2ST data only yields up to 1 BLEU gain over the baseline \cite{enhanced_s2u_popuri2022}. However, adding the \textaug data offers an extra 1 BLEU improvement.

\vspace{-0.8em}
\section{Conclusion}
\vspace{-0.2em}
We presented an effective method to generate a large amount of synthetic S2ST data from unlabeled text corpora, as well as an online audio augmentation process that can transform them into more prosodically diverse, which is shown to outperform the state of the art in the speech-to-speech Es-En translation tasks by up to 2 ASR-BLEU. In our analysis, we also demonstrate that the method is also more beneficial for low-resource and unsupervised setups.

\vspace{-0.8em}
\section{Acknowledgement}
We would like to thank Ann Lee for technical discussions and Peng-Jen Chen for his help with TTS models.

    % \section{RELATION TO PRIOR WORK}
    % \label{sec:prior}
    
    % The text of the paper should contain discussions on how the paper's
    % contributions are related to prior work in the field. It is important
    % to put new work in  context, to give credit to foundational work, and
    % to provide details associated with the previous work that have appeared
    % in the literature. This discussion may be a separate, numbered section
    % or it may appear elsewhere in the body of the manuscript, but it must
    % be present.
    
    % You should differentiate what is new and how your work expands on
    % or takes a different path from the prior studies. An example might
    % read something to the effect: "The work presented here has focused
    % on the formulation of the ABC algorithm, which takes advantage of
    % non-uniform time-frequency domain analysis of data. The work by
    % Smith and Cohen \cite{Lamp86} considers only fixed time-domain analysis and
    % the work by Jones et al \cite{C2} takes a different approach based on
    % fixed frequency partitioning. While the present study is related
    % to recent approaches in time-frequency analysis [3-5], it capitalizes
    % on a new feature space, which was not considered in these earlier
    % studies."

\vfill\pagebreak

    % \section{REFERENCES}
    % \label{sec:refs}
    
    % List and number all bibliographical references at the end of the
    % paper. The references can be numbered in alphabetic order or in
    % order of appearance in the document. When referring to them in
    % the text, type the corresponding reference number in square
    % brackets as shown at the end of this sentence \cite{C2}. An
    % additional final page (the fifth page, in most cases) is
    % allowed, but must contain only references to the prior
    % literature.

% \bibliographystyle{IEEEbib}
\bibliographystyle{IEEEbibShort}
\bibliography{refs_short}

\begin{thebibliography}{10}

\bibitem{self_train_asr_kahn2020}
J.Kahn, A.Lee, et~al.,
\newblock ``Self-training for end-to-end speech recognition,''
\newblock ICASSP. 2020.

\bibitem{vaswani2017attention}
A.Vaswani, N.Shazeer, et~al.,
\newblock ``Attention is all you need,''
\newblock NeurIPS, 2017.

\bibitem{criss2020}
C.Tran, Y.Tang, et~al.,
\newblock ``Cross-lingual retrieval for iterative self-supervised training,''
\newblock NeurIPS. 2020.

\bibitem{swavumt2021}
X.-P.Nguyen, H.Gong, et~al.,
\newblock ``Contrastive clustering to mine pseudo parallel data for
  unsupervised translation,''
\newblock ICLR, 2022.

\bibitem{s2t_bansal2017towards}
S.Bansal, H.Kamper, et~al.,
\newblock ``Towards speech-to-text translation without speech recognition,''
\newblock {\em arXiv:1702.03856}, 2017.

\bibitem{vits_kim2021conditional}
J.Kim, J.Kong, et~al.,
\newblock ``Conditional variational autoencoder with adversarial learning for
  end-to-end text-to-speech,''
\newblock ICML. 2021.

\bibitem{fastspeech_ren2019}
Y.Ren, Y.Ruan, et~al.,
\newblock ``Fastspeech: Fast, robust and controllable text to speech,''
\newblock {\em NeurIPS}, 2019.

\bibitem{janus_iii_lavie1997janus}
A.Lavie, A.Waibel, et~al.,
\newblock ``Janus-iii: Speech-to-speech translation in multiple languages,''
\newblock ICASSP. 1997.

\bibitem{ATR_speech_to_speech_1597243}
S.Nakamura, K.Markov, et~al.,
\newblock ``The atr multilingual speech-to-speech translation system,''
\newblock {\em ICASSP}, 2006.

\bibitem{direct_s2s_jia2019}
Y.Jia, R.~J.Weiss, et~al.,
\newblock ``Direct speech-to-speech translation with a sequence-to-sequence
  model,''
\newblock {\em arXiv:1904.06037}, 2019.

\bibitem{s2s_discrete_units_lee2021}
A.Lee, P.-J.Chen, et~al.,
\newblock ``Direct speech-to-speech translation with discrete units,''
\newblock {\em arXiv:2107.05604}, 2021.

\bibitem{enhanced_s2u_popuri2022}
S.Popuri, P.-J.Chen, et~al.,
\newblock ``Enhanced direct speech-to-speech translation using self-supervised
  pre-training and data augmentation,''
\newblock {\em arXiv:2204.02967}, 2022.

\bibitem{textless-s2s-lee-etal-2022}
A.Lee, H.Gong, et~al.,
\newblock ``Textless speech-to-speech translation on real data,''
\newblock NAACL, Seattle, United States, July 2022.

\bibitem{wav2vec2_baevski2020}
A.Baevski, Y.Zhou, et~al.,
\newblock ``wav2vec 2.0: A framework for self-supervised learning of speech
  representations,''
\newblock {\em NeurIPS}, 2020.

\bibitem{lakhotia2021generative}
K.Lakhotia, E.Kharitonov, et~al.,
\newblock ``On generative spoken language modeling from raw audio,''
\newblock {\em ACL}, 2021.

\bibitem{discrete_unit_polyak2021speech}
A.Polyak, Y.Adi, et~al.,
\newblock ``Speech resynthesis from discrete disentangled self-supervised
  representations,''
\newblock {\em arXiv:2104.00355}, 2021.

\bibitem{mbart_liu2020multilingual}
Y.Liu, J.Gu, et~al.,
\newblock ``Multilingual denoising pre-training for neural machine
  translation,''
\newblock {\em ACL}, 2020.

\bibitem{weaklysup_s2t_jia2019leveraging}
Y.Jia, M.Johnson, et~al.,
\newblock ``Leveraging weakly supervised data to improve end-to-end
  speech-to-text translation,''
\newblock ICASSP. 2019.

\bibitem{ccnet_data_wenzek2019}
G.Wenzek, M.-A.Lachaux, et~al.,
\newblock ``Ccnet: Extracting high quality monolingual datasets from web crawl
  data,''
\newblock {\em arXiv:1911.00359}, 2019.

\bibitem{xlm_conneau2019cross}
A.Conneau and G.Lample,
\newblock ``Cross-lingual language model pretraining,''
\newblock {\em NeurIPS}, 2019.

\bibitem{hubert_hsu2021}
W.-N.Hsu, B.Bolte, et~al.,
\newblock ``Hubert: Self-supervised speech representation learning by masked
  prediction of hidden units,''
\newblock {\em ICASSP}, 2021.

\bibitem{s2t_self_learn_wang2021large}
C.Wang, A.Wu, et~al.,
\newblock ``Large-scale self-and semi-supervised learning for speech
  translation,''
\newblock {\em arXiv:2104.06678}, 2021.

\bibitem{hifigan_vocoder_polyak2021}
A.Polyak, Y.Adi, et~al.,
\newblock ``Speech resynthesis from discrete disentangled self-supervised
  representations,''
\newblock {\em arXiv:2104.00355}, 2021.

\bibitem{conform_gulati2020}
A.Gulati, J.Qin, et~al.,
\newblock ``Conformer: Convolution-augmented transformer for speech
  recognition,''
\newblock {\em arXiv:2005.08100}, 2020.

\bibitem{layernorm_ba2016layer}
J.~L.Ba, J.~R.Kiros, et~al.,
\newblock ``Layer normalization,''
\newblock {\em arXiv:1607.06450}, 2016.

\bibitem{musan2015}
D.Snyder, G.Chen, et~al.,
\newblock ``{MUSAN}: {A} {M}usic, {S}peech, and {N}oise {C}orpus,'' 2015,
\newblock arXiv:1510.08484v1.

\bibitem{snr_plapous2006improved}
C.Plapous, C.Marro, et~al.,
\newblock ``Improved signal-to-noise ratio estimation for speech enhancement,''
\newblock {\em ICASSP}, 2006.

\bibitem{understand_bt_edunov2018}
S.Edunov, M.Ott, et~al.,
\newblock ``Understanding back-translation at scale,''
\newblock {\em arXiv:1808.09381}, 2018.

\bibitem{covost2_wang2020covost}
C.Wang, A.Wu, et~al.,
\newblock ``Covost 2 and massively multilingual speech-to-text translation,''
\newblock {\em arXiv:2007.10310}, 2020.

\bibitem{europarl_iranzo2020}
J.Iranzo-S{\'a}nchez, J.~A.Silvestre-Cerda, et~al.,
\newblock ``Europarl-st: A multilingual corpus for speech translation of
  parliamentary debates,''
\newblock ICASSP. 2020.

\bibitem{mtedx_elizabeth2021multilingual}
S.Elizabeth, W.Matthew, et~al.,
\newblock ``The multilingual tedx corpus for speech recognition and
  translation,''
\newblock Interspeech, 2021.

\bibitem{mls_pratap2020mls}
V.Pratap, Q.Xu, et~al.,
\newblock ``Mls: A large-scale multilingual dataset for speech research,''
\newblock {\em arXiv preprint arXiv:2012.03411}, 2020.

\bibitem{common_voice_ardila2019}
R.Ardila, M.Branson, et~al.,
\newblock ``Common voice: A massively-multilingual speech corpus,''
\newblock {\em arXiv preprint arXiv:1912.06670}, 2019.

\bibitem{libri_kahn2020libri}
J.Kahn, M.Rivi{\`e}re, et~al.,
\newblock ``Libri-light: A benchmark for asr with limited or no supervision,''
\newblock ICASSP. 2020.

\bibitem{tedlium_hernandez2018ted}
F.Hernandez, V.Nguyen, et~al.,
\newblock ``Ted-lium 3: twice as much data and corpus repartition for
  experiments on speaker adaptation,''
\newblock SPECOM. 2018.

\bibitem{voxpopuli_wang2021}
C.Wang, M.Riviere, et~al.,
\newblock ``Voxpopuli: A large-scale multilingual speech corpus for
  representation learning, semi-supervised learning and interpretation,''
\newblock {\em arXiv:2101.00390}, 2021.

\end{thebibliography}

\end{document}